\pdfoutput=1

\documentclass[a4]{article}

\usepackage{standalone}
\usepackage{latexsym}
\usepackage{soul,color,xcolor}

\usepackage[T1]{fontenc}

\usepackage[utf8]{inputenc}
\usepackage{graphicx}
\usepackage{arydshln}
\usepackage{pifont}

\usepackage{microtype}

\usepackage{inconsolata}

\usepackage{natbib}
\usepackage[colorlinks, citecolor=black, linkcolor=black]{hyperref}
\newcommand\figwidth{0.5\textwidth}
\newcommand\toc1

\author{
  \textbf{Swaroop Mishra}\thanks{
    Equal first authors. 
  }~\thanks{Work done while at the Allen Institute for AI.}
  \\ Arizona State University
  \and \textbf{Matthew Finlayson}\footnotemark[1]~\thanks{Corresponding authors:
    \href{matthewf@allenai.org}{matthewf@allenai.org},
    \href{ashwinkv@allenai.org}{ashwinkv@allenai.org}.} \\ The Allen Institute for AI 
  \and \textbf{Pan Lu}\footnotemark[2] \\ UCLA
  \and \textbf{Leonard Tang} \\ Harvard University
  \and \textbf{Sean Welleck} \\ The Allen Institute for AI 
  \and \textbf{Chitta Baral} \\ Arizona State University
  \and \textbf{Tanmay Rajpurohit} \\ Georgia Institute of Technology
  \and \textbf{Oyvind Tafjord} \\ The Allen Institute for AI
  \and \textbf{Ashish Sabharwal} \\ The Allen Institute for AI
  \and \textbf{Peter Clark} \\ The Allen Institute for AI
  \and \textbf{Ashwin Kalyan}\footnotemark[3] \\ The Allen Institute for AI
} 

\date{}

\usepackage{multirow}
\usepackage{booktabs}
\usepackage{subcaption}
\usepackage{fontawesome5}

\usepackage{xspace}
\usepackage{boxedminipage}
\usepackage{amsmath,amssymb}
\usepackage{enumitem}

\newcommand{\lila}{\textsc{L\=ila}\xspace}
\newcommand{\lilatest}{\textsc{L\=ila-Test}\xspace}
\newcommand{\lilarobust}{\textsc{L\=ila-Robust}\xspace}
\newcommand{\lilaood}{\textsc{L\=ila-OOD}\xspace}
\newcommand{\bhaskara}{\textsc{Bh\=askara}\xspace}

\newcommand{\model}{\textsc{Bh\=askara}\xspace}
\newcommand{\dsl}[1]{\texttt{#1}\xspace}

\newcommand{\mf}[1]{#1}

\newcommand{\eg}{e.g.,\xspace}
\newcommand{\ie}{i.e.,\xspace}
\newcommand{\vs}{vs.\xspace}
\newcommand{\cmark}{\textcolor{green}{\ding{51}}}%
\newcommand{\xmark}{\textcolor{red}{\ding{55}}}%

\usepackage[most]{tcolorbox}
\usepackage{pythonhighlight}
\usepackage{pifont}
\usepackage{multicol}
\usepackage[framemethod=tikz]{mdframed}
\definecolor{qualcolor}{RGB}{128,64,0}
\gdef\Sepline{%
  \par\noindent\makebox[\linewidth][l]{%
  \hspace*{-\mdflength{innerleftmargin}}%
   \tikz\draw[thick,dashed,gray!60] (0,0) --%
        (\textwidth+\the\mdflength{innerleftmargin}+\the\mdflength{innerrightmargin},0);
  }\par\nobreak}

\title{\lila: A Unified Benchmark for Mathematical Reasoning}

\begin{document}

\maketitle


\begin{abstract}
  \mf{
    Mathematical reasoning skills are essential for general-purpose intelligent
    systems to perform tasks from grocery shopping to climate modeling.
  }
  Towards evaluating and improving AI systems in this domain, we propose
  \lila, a unified mathematical reasoning benchmark consisting of 23 diverse
  tasks along four dimensions:
  (i) mathematical abilities \eg arithmetic, calculus 
  (ii) language format \eg question-answering, fill-in-the-blanks 
  (iii) language diversity \eg no language, simple language 
  (iv) external knowledge \eg commonsense, physics. 
  \mf{
    We construct our benchmark by extending 20 datasets benchmark 
    by collecting task instructions and solutions in the form of Python programs,
    thereby obtaining explainable solutions 
    in addition to the correct answer.
  }
  We \mf{additionally} introduce two evaluation datasets 
  to measure out-of-distribution performance and robustness to language perturbation.
  Finally, we introduce \model,
  a general-purpose mathematical reasoning model trained on \lila. 
  Importantly, we find that multi-tasking leads to significant improvements 
  (average relative improvement of $21.83\%$ F1 score \vs single-task models),
  while the best performing model only obtains $60.40\%$,
  indicating the room for improvement 
  in general mathematical reasoning and understanding.\footnotemark
\end{abstract}

\footnotetext{Our dataset: \url{https://github.com/allenai/Lila}.
Our model: \url{https://huggingface.co/allenai/bhaskara}.}


\section{Introduction}

\begin{figure}[t]
  \centering
 \includegraphics[width=\figwidth]{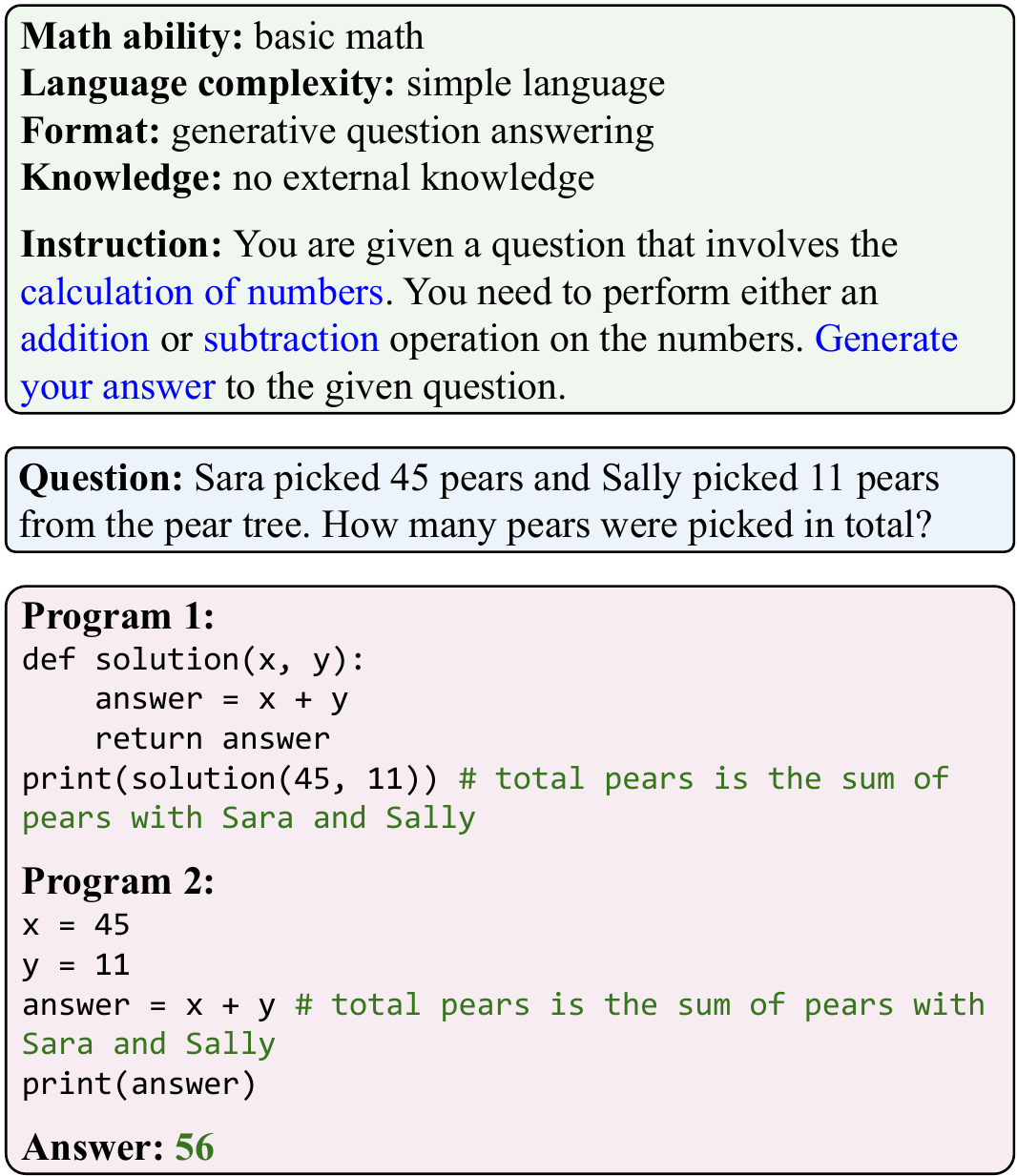}
  \caption{A data example with two Python programs in \lila. 
  One program annotation uses function construct whereas the other one is a plain script without function. The instruction for each task and categories across four dimensions are annotated for developing \lila{}.}
  \label{fig:example}
\end{figure}

Mathematical reasoning is required in all aspects of life,
from buying \mf{ingredients for a recipe} to controlling the world economy. 
Given the fundamental nature of mathematical reasoning, a \mf{number} of works 
propose datasets to evaluate \mf{specific} mathematical reasoning abilities of AI agents,
\eg \citet{kushman2014learning} (algebra word problems),
\citet{mishra2022numglue} (arithmetic reasoning),
\citet{saxton2019analysing} (templated math reasoning spanning algebra, calculus, probability, etc.)
\mf{
  Since evaluating high-capacity models 
  on narrowly scoped mathematical reasoning datasets 
  risks overestimating the reasoning abilities of these AI systems,
  creating the need for a unified benchmark for systematic evaluation over 
  diverse topics and problem styles. 
}

To this end, we introduce \lila\footnotemark,
a unified mathematical reasoning benchmark that consists of 23  mathematical reasoning tasks.
\footnotetext{
  Named after \textit{L\=ilavati}, a 12\textsuperscript{th} century
  mathematical treatise on arithmetic that covers topics like arithmetic and
  geometric progressions, indeterminate equations and combinations. It is also
  widely known for the extensive number of math word problems. The author,
  \textit{Bh\=askara} is known for fundamental and original contributions to
  calculus, physics, number theory, algebra, 
  and astronomy~\citep{colebrooke1817arithmetic, sarkar1918hindu,  kolachana2019use}
}
\lila is constructed by extending 20 existing datasets 
\mf{spanning a wide range of topics in mathematics,}
\mf{varying degrees of linguistic complexity,}
\mf{and diverse question formats and background knowledge requirements.}
Importantly, \lila extends all of these datasets to include a solution program
\mf{
  as opposed to only an answer, 
  and instruction annotations 
  to enable  instruction-based learning~\citep{sanh2021multitask,wei2021finetuned, mishra2022cross}.
}

\mf{
  In order to accurately assess the mathematical reasoning ability of models,
  evaluating the chain of reasoning that leads to the correct solution is equally
  important (if not more important) to evaluating the final answer or
  expression. 
}
We therefore collect Python programs that serve as \mf{reasoning chains} for each question in the benchmark. 
\mf{
  We achieve this by automatically converting domain-specific language (DSL) annotations 
  into Python programs 
  and by manually collecting expert annotations 
  when no DSL annotations are available.
}
\mf{By incorporating program annotations,} 
\lila unifies various mathematical reasoning datasets under a single problem formulation 
\ie given an input problem in natural language,
generate a Python program that upon execution returns the desired answer. 
\mf{This formulation} allows neural approaches to focus on the high-level aspects of
mathematical problem solving 
(\eg identifying potential solution strategies, decomposing the problem into simpler sub-problems),
while leveraging external solvers 
(\eg Python builtins, Sympy) 
to perform precise operations like adding huge numbers or simplifying expressions. 
Figure \ref{fig:example} illustrates a sample from our \lila benchmark that
illustrates the question, answer, program, instructions, and category tags.

\mf{In addition to evaluating high-level problem solving,}
we also facilitate two other key ways to make a fair assessment of models on mathematical reasoning tasks. 
In line with \citet{bras2020adversarial}, \citet{ ribeiro2020beyond} and \citet{welleck2022symbolic}, we evaluate generalization 
\eg alternate formulations of a problem (``2+2=?'' \vs ``What is two plus two?'') 
using an out-of-distribution evaluation set (\lilaood) 
containing datasets requiring the same underlying mathematical reasoning skills,
but were collected independently of the training datasets. 
Further, we collect a robustness split \lilarobust, 
that introduces linguistic perturbations (\eg active \vs passive voice) via crowd-sourcing. 
The evaluation scheme is a combination of the performance on all three sets: \lilatest, \lilaood and \lilarobust.

\paragraph{Contributions}
\begin{enumerate}[leftmargin=15pt, noitemsep, nolistsep]
  \item 
    \mf{
      We present \lila,
      a holistic benchmark for mathematical reasoning.
      \lila extends 20 existing datasets 
      with solutions in the form of Python programs and instruction annotations, 
      and categorizes questions into 23 tasks based on their
      language complexity, question format and need for external knowledge. 
      Our benchmark measures performance on out-of-distribution examples 
      and robustness to language perturbations 
      in addition to standard test-set. 
    }
  \item We introduce \model, a multi-task model fine-tuned on our dataset.
    Our best-performing model achieves comparable performance to a 66$\times$
    larger model pre-trained on both code and language.
  \item 
    We provide an analysis of our models' performance and find that 
    (1) multitasking improves considerably over task-specific learning 
    both in in-distribution and out-of-distribution evaluation 
    (2) program synthesis substantially outperforms answer prediction, 
    (3) few-shot prompting with codex has the strongest performance.
    \mf{
      We also identify areas for improvement for future work, 
      e.g., data gaps in \lila categories.
    }
\end{enumerate}

\section{Related Work}

\paragraph{Mathematical Reasoning Datasets.}
Our work builds on an existing body of mathematical reasoning literature. 
Early work in this areas focuses on small-scale datasets testing
addition-subtraction~\citep{Hosseini14learningto}, templated questions with
equations as parameters~\citep{kushman2014learning} and other forms of
arithmetic reasoning~\citep{koncel2015parsing, roy2016solving, upadhyay2016learning, roy2017unit, roy2018mapping, ling2017program}. 
Later datasets increase in complexity and scale, 
incorporating reading comprehension \cite{dua2019drop},  algebra~\citep{saxton2019analysing}, and multi-modal contexts~\citep{lu2021inter,lu2022dynamic}.
Still other numerical reasoning datasets focus on diversity~\citep{miao-etal-2020-diverse} 
with multiple categories of numerical reasoning tasks~\citep[e.g.,][]{amini2019mathqa}. 
Most recently, new datasets have focused on increasing difficulty, 
e.g., olympiad problems~\citep{hendrycks2021measuring} 
and adversarial problems~\citep{patel_etal_2021_nlp}, 
as well as increasing the knowledge requirements to solve tasks, 
with a growing focus on commonsense reasoning~\citep{zhou2019going, zhanglanguage, lu2021iconqa, mishra2022numglue}. 

A separate line of work in mathematical reasoning 
includes datasets testing mathematical theorem
proving~\citep[e.g.,][]{li2021isarstep,wu2021int,welleck2021naturalproofs,zheng2021minif2f,
Han2021ProofAC}.
We do not, however, consider theorem proving in our work, choosing instead to focus on numerical reasoning.

\paragraph{Task Hierarchy and Multi-tasking in Numerical Reasoning.}
We take inspiration from the success of multi-task learning in NLP~\citep{weston2015towards},
including benchmarks~\citep[e.g.,][]{wang2018glue, wang2019superglue, dua2019orb} 
and multitasking models~\citep[e.g.,][]{McCann2018decaNLP, khashabi2020unifiedqa, lourie2021unicorn, aghajanyan2021muppet}.
NumGLUE~\citep{mishra2022numglue} has been proposed as a multi-tasking numerical reasoning benchmark that contains 8 different tasks. 
\lila expands NumGLUE to provide wider coverage of mathematical abilities,
along with evaluation that captures out-of-domain, robustness, and instruction-following performance.
Our introduction of mathematical reasoning categories and the evaluation setup
is inspired by task hierarchies in other domains 
such as vision~\citep{zamir2018taskonomy} and NLP~\citep{rogers2021qa} 
which appear in large scale benchmarks~\citep[e.g.,][]{srivastava2022beyond, wang2022benchmarking}.

\section{\lila}

\lila is composed of 23 tasks across 4 dimensions, curated from 44 sub-datasets across 20 dataset sources. 
Here we discuss the construction and composition of the benchmark 
and provide descriptive statistics of the datasets. 

\begin{table*}
  \centering
  \small
  \begin{tabular}{lp{0.78\textwidth}}
  \toprule
  Category & Tasks \\
  \midrule
  Math ability & Basic math, multiplication/division, number theory, algebra, geometry, counting and statistics, calculus, linear algebra, advanced math \\
  Language     & No language, simple language, complex language \\
  Knowledge    & No background knowledge, commonsense, math, science, computer science, real world knowledge \\
  Format       & Fill-in-the-blank, generative question answering, multiple-choice, natural language inference, reading comprehension \\
  \bottomrule
  \end{tabular}
  \caption{Categories and their associated tasks.}\label{tab:tasks}
\end{table*}

\subsection{Dataset Construction}

\paragraph{Data Sources.} 
\lila incorporates 20 existing datasets from the mathematical reasoning literature (Table~\ref{tab:source datasets} gives a detailed list),
where inputs are natural language or templated text 
and outputs are numerical or expressions, 
\eg we exclude theorem proving~\citep{welleck2021naturalproofs, Han2021ProofAC},
where the output is not a number or expression.
We leave the incorporation of formats like theorem proving to future work.

\paragraph{Unified format.} 
We normalize all datasets to a unified format with the following fields:
\begin{enumerate}[leftmargin=15pt, noitemsep, nolistsep]
  \item The source dataset.
Category tags for each of the four dimensions
    (math ability, language complexity, format, and external knowledge; see \S\ref{sec:cat}). 
  \item The question, in English.
  \item The answer to the question, as a string containing a number, expression, list, or other data format.
A set of Python strings that \texttt{print} the answer.
  \item A task-level instruction in natural language.
\end{enumerate}
We also retain meta-data from the original dataset.

\paragraph{Automatic program annotation.}
Most of the annotations in the source datasets do not contain output in the form of a Python program. 
We automatically annotate most datasets by generating Python programs using the
annotations (answer, explanation, etc.) 
provided in the source datasets.
Where possible, we generate multiple Python programs for a single question. 
This is to account for variation in the program space 
such as the choice of data structure, language construct, variable name, and programming style 
(e.g., declarative vs procedural).
For example, Figure~\ref{fig:example} gives multiple Python programs solving the same question;
in this case one program directly calculates the answer, 
whereas the other defines a function to solve the problem more generally.

Some datasets contain program annotations that can be captured by a domain-specifc language (DSL) 
in which case we write rules to convert them into Python programs,
\eg \dsl{volume(sphere,3)} to the Python expression \texttt{4/3*math.pi*3**3}.
In some cases where a DSL annotation is not provided,
we use pattern matching to convert highly templated datasets 
like the AMPS dataset~\citep{hendrycks2021measuring} to our unified format.
In other cases, instead of converting the existing dataset,
we modify the data generation code to reproduce the dataset with program annotations. 
For the DeepMind mathematics dataset \citep{saxton2019analysing}, 
this allows us to create diverse, compositional math problems with program annotations using a sophisticated grammar.

\paragraph{Expert program annotation.}
For many datasets, it is not possible to obtain Python program annotations
via automated methods described above;
either the original dataset contains only the final answer 
or contains solutions expressed in free-form natural language. 
For such datasets, we obtain annotations from experts who are proficient in basic programming and high-school level mathematics. 
See Appendix~\ref{app:expert} for details.

\paragraph{Instruction annotation.}
Given the effectiveness of instruction learning~\citep{mishra2022cross, wei2021finetuned, mishra-etal-2022-reframing, sanh2021multitask} for effective generalization,
we collect instruction annotation for each task.
Each instruction contains a \textit{definition} that clearly defines the task and provides guidelines,
a \textit{prompt} that provides a short and straight forward instruction,
and \textit{examples} that facilitate learning by demonstration~\citep{brown2020language}.
Figure~\ref{fig:example} shows an example instruction for the basic math task (\S \ref{sec:cat}).

\subsection{Categories and Tasks} 
\label{sec:cat}
We create 4 \emph{views}\footnotemark or categories of \lila 
along the dimensions of mathematical area, language complexity, external knowledge, and question format. 
\footnotetext{Note that it is \emph{not} a partition of the benchmark as each dimensions divides the constituent examples in different ways} 
Altogether, these views classify the data into 23 \emph{tasks}
(Table~\ref{tab:tasks}). 
By creating multiple views of the benchmark,
we are able to systematically characterize the strengths and weaknesses of existing models 
at a granular level.

The first category, \emph{math ability}, partitions the datasets into common pedagogical subjects:
arithmetic, algebra, geometry, calculus, etc. 

Our second category, \emph{language complexity},
separates math problems by the complexity of the language used to represent them.
This ranges from formal representations only (e.g., \texttt{1+1=?}) 
to natural language (e.g., ``Mariella has 3 pears\ldots''). 

We next partition datasets based on the type of \emph{background knowledge},
required to solve the problem. 
For instance, commonsense questions like ``How many legs to 3 people have?'' 
or science questions like ``Will water boil at 200 degrees Celsius?'' 
require different sets of knowledge to answer. 

Lastly, we categorize based on \emph{question format}, 
putting e.g., multiple choice questions under one task and natural language inference under another. 
Examples of each task and the datasets included are in Appendix~\ref{app:task-desc}.

\subsection{\lilaood}
In order to measure if the model has truly learned the underlying mathematical reasoning skill,
we evaluate both in-distribution (IID, i.e., standard train-test splits) 
and out-of-distribution (OOD) performance for each task,
\ie we evaluate on examples requiring the \emph{same} underlying mathematical reasoning skill
but from a different dataset.
To construct \lilaood, we follow \citet{bras2020adversarial} and \citet{ hendrycks2020pretrained}
by randomly assigning the datasets for each task into IID and an OOD sets,
using the IID set for training and standard evaluation
and the OOD set to evaluate generalization.
We do not include tasks in \lilaood for tasks containing only one dataset.

\subsection{\lilarobust}
\label{robust}
In light of recent work demonstrating the brittleness of language models 
at solving math problems~\citep{patel_etal_2021_nlp},
we create a high-quality evaluation dataset, \lilarobust,
to evaluate performance on mathematical reasoning tasks 
when linguistic perturbations are introduced.
Specifically, we define and apply a set of carefully chosen augmentation templates,
summarized in Table \ref{tab:robust-examples},
on each task,
yielding a set of challenging problems that are consistent answer-wise 
but stylistically different question-wise. 
Overall, we define a total of 9 templates for such question perturbations: 
3 from \citet{patel_etal_2021_nlp} and 6 of our own. 
From each constituent dataset,
we sample 20 questions and obtain perturbed question annotations via Amazon Mechanical Turk (AMT). 
Refer to Appendix \ref{robust-collection} for additional details on the construction of \lilarobust.




\subsection{Statistics}

Table \ref{tab:stat} shows key statistics of our proposed benchmark, \lila. 
\lila contains $\approx134$K examples with significant diversity across question,
answer, program and instruction length 
(see detailed statistics in Appendix~\ref{app:data_stats}). 
Figure ~\ref{fig:question_ngram} shows the diversity of
questions in \lila. Note that we downsample (via random selection) some
datasets like AMPS~\citep{hendrycks2021measuring} which contains numerous templated questions that can get over-representated in the distribution of examples across categories in \lila{}.

\begin{table}[t]
\centering
\small
\begin{tabular}{lc}
\toprule
\textbf{Statistic}  & \textbf{Number}  \\ 
\midrule
\# Total tasks & 23 \\
\# Total datasets & 44 \\
\# Total instructions & 44 \\
\# Total questions & 133,815 \\
\# Total programs & 358,769  \\
\midrule
Unique questions & 132,239  \\
Unique programs & 325,597  \\
Unique answers & 271,264  \\
\midrule
Average length of instructions &31.18  \\
Average length of questions & 47.72 \\
Average length of programs  & 47.85 \\
\bottomrule
\end{tabular}
\caption{Key statistics of \lila.}
\label{tab:stat}
\end{table}

\begin{figure}
  \centering
  \includegraphics[width=\figwidth]{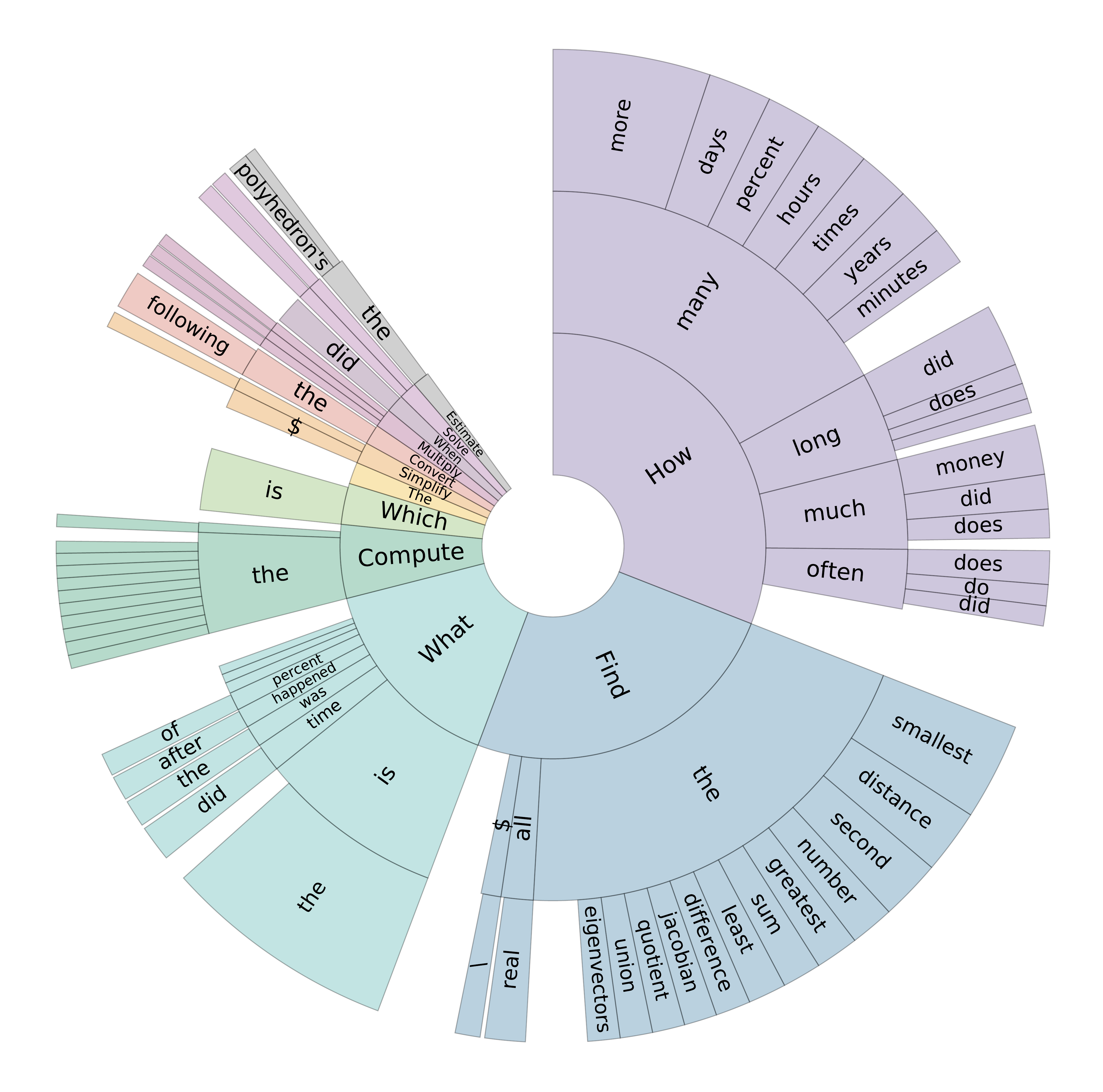}
  \caption{Question n-gram distribution in \lila.}
  \label{fig:question_ngram}
\end{figure}

\section{Experiments}
In this section, we introduce our modeling contributions for the \lila benchmark and discuss the overall experimental setup.


\paragraph{Data partition and evaluation.} 
For the IID setup, we randomly partition the data in \emph{each} task into training (70\%), development (10\%) and test (20\%) sets. 
Additionally, we also evaluate on \lilaood and \lilarobust settings; thus, the
final evaluation scheme is a combination of the performance on all three
evaluation setups

\paragraph{Fine-tuning.}
We fine-tune a series of GPT-Neo-2.7B causal language models~\citep{gpt-neo}) on \lila.
We choose GPT-Neo because it was pre-trained 
on both natural language and code~\citep{pile}, 
as opposed to solely on natural language.
To assess the capabilities of GPT-Neo 
on various aspects of the dataset,
we fine-tune \emph{single-task} models on each of the 23 tasks in \lila.
We also evaluate the benefit of transfer learning
by fine-tuning a single \emph{multi-task} GPT-Neo baseline 
on all the tasks simultaneously.
We call our multitask model \bhaskara.

\paragraph{Prompting.}
We also use few-shot prompting 
to evaluate GPT-3 
and Codex\footnote{\texttt{text-davinci-002, code-davinci-002}}~\citep{brown2020language, chen2021evaluating}.
For the IID setting,
we prompt the model with a random input-output examples from the same dataset as the input.
In the OOD setting, we take examples from other datasets (Table~\ref{tab:data_source_math}-\ref{tab:data_source_knowledge}) within the same task.
We repeat this evaluation with increasing numbers of examples (up to the token
size of models) to study the effect on performance\footnote{Henceforth we
refer to the max example model unless otherwise specified.}.

\paragraph{Evaluation.} We evaluate our models under two regimes---directly outputting the answer \ie program induction and outputting a Python program that is then executed to obtain the final answer \ie program synthesis.
In the case of our fine-tuned models, we train them to output both the final answer and the Python program conditioned on the input question. 
To evaluate our models under direct question answering, 
we use F1-score\footnote{This is a soft version of exact match accuracy assigning partial credit when common words are present in the output and gold answer.} to compare the model output and the gold answer. 
To evaluate program synthesis, 
we execute the model's output within a Python interpreter
and compare the program output with the output of the gold program, 
again using F1.
We evaluate based on the program output, rather than the program itself,
to account for diversity in solving techniques and programming styles.

\begin{table*}[t]
\centering
\renewcommand\tabcolsep{5.0pt} 
\resizebox{1.0\linewidth}{!}{
\begin{tabular}{clcc|cc||cc|cc|cc|cc} 
\toprule
& $\rightarrow$ {\textbf{Supervision/Size}} & \multicolumn{2}{c|}{Few-shot, 175B} & \multicolumn{2}{c||}{Few-shot, 175B} & \multicolumn{2}{c|}{Fine-tuned, 2.7B} & \multicolumn{2}{c|}{Fine-tuned, 2.7B} & \multicolumn{2}{c|}{Fine-tuned, 2.7B} & \multicolumn{2}{c}{Fine-tuned, 2.7B}\\
\midrule
\multirow{2}{*}{$\downarrow$ \textbf{Task}} & \multirow{2}{*}{\textbf{Category}} & \multicolumn{2}{c|}{\textbf{GPT-3}} & \multicolumn{2}{c||}{\textbf{Codex}} & \multicolumn{2}{c|}{\textbf{Neo-A}} & \multicolumn{2}{c|}{\textbf{Neo-P}} & \multicolumn{2}{c|}{\textbf{\bhaskara-A}} & \multicolumn{2}{c}{\textbf{\bhaskara-P}}  \\
 & & IID & OOD & IID & OOD & IID & OOD & IID & OOD & IID & OOD & IID & OOD  \\ 
\midrule
1 & Basic math & 0.766 & \underline{\textbf{0.818}} & \underline{\textbf{0.791}} & 0.762 & 0.533 & 0.523 & 0.611 & 0.555 & 0.693 & 0.657 & \textbf{0.790} & \textbf{0.787} \\
2 & Muldiv & 0.479 & 0.665 & \underline{\textbf{0.691}} & \underline{\textbf{0.790}} & 0.136 & 0.089 & 0.388 & 0.194 & 0.155 & 0.083 & \textbf{0.448} & \textbf{0.395} \\
3 & Number theory & 0.240 & 0.154 & \underline{\textbf{0.472}} & \underline{\textbf{0.344}} & 0.108 & 0.095 & 0.328 & 0.107 & 0.129 & 0.190 & \textbf{0.358} & \textbf{0.293} \\
4 & Algebra & 0.338 & 0.130 & \underline{\textbf{0.603}} & \underline{\textbf{0.511}} & 0.164 & 0.031 & 0.348 & 0.051 & 0.203 & \textbf{0.054} & \textbf{0.473} & 0.007 \\
5 & Geometry & \textbf{0.283} & 0.120 & 0.000 & \underline{\textbf{\textbf{0.250}}} & 0.288 & 0.025 & 0.077 & 0.021 & \underline{\textbf{0.297}} & 0.105 & 0.079 & \underline{\textbf{\textbf{0.250}}} \\
6 & Statistics & 0.183 & \underline{\textbf{0.210}} & \textbf{0.650} & 0.200 & 0.107 & 0.008 & 0.839 & 0.034 & 0.115 & \textbf{0.179} & \underline{\textbf{0.947}} & 0.164 \\
7 & Calculus & 0.231 & 0.208 & \underline{\textbf{0.930}} & \underline{\textbf{0.884}} & 0.138 & 0.119 & 0.486 & 0.334 & 0.102 & 0.167 & \textbf{0.495} & \textbf{0.805} \\
8 & Linear algebra & 0.127 & - & \textbf{0.692} & - & 0.229 & - & \underline{\textbf{0.809}} & - & 0.240 & - & 0.808 & - \\
9 & Advanced math & 0.150 & - & \underline{\textbf{0.472}} & - & 0.012 & - & 0.100 & - & 0.019 & - & \textbf{0.160} & - \\
\midrule
10 & No language & 0.213 & 0.162 & \underline{\textbf{0.853}} & \textbf{0.770} & 0.143 & 0.083 & 0.698 & 0.330 & 0.140 & 0.138 & \textbf{0.703} & \underline{\textbf{0.850}} \\
11 & Simple language & 0.486 & 0.561 & \underline{\textbf{0.568}} & \underline{\textbf{0.610}} & 0.269 & 0.243 & 0.363 & 0.292 & 0.332 & 0.269 & \textbf{0.433} & \textbf{0.384} \\
12 & Complex language & 0.356 & 0.413 & \underline{\textbf{0.456}} & \underline{\textbf{0.583}} & 0.147 & 0.113 & 0.216 & 0.106 & 0.215 & 0.259 & \textbf{0.288} & \textbf{0.557} \\
\midrule
13 & Fill in the blank & 0.710 & 0.620 & \underline{\textbf{0.790}} & \underline{\textbf{0.660}} & 0.086 & 0.193 & \textbf{0.304} & 0.193 & 0.059 & \textbf{0.519} & 0.262 & \textbf{0.519} \\
14 & Generative QA & 0.305 & 0.385 & \underline{\textbf{0.566}} & \underline{\textbf{0.632}} & 0.142 & 0.135 & 0.376 & 0.199 & 0.178 & 0.160 & \textbf{0.476} & \textbf{0.235} \\
15 & MCQ & \textbf{0.801} & \textbf{0.870} & 0.771 & \textbf{0.870} & 0.636 & 0.818 & 0.652 & 0.818 & 0.752 & \underline{\textbf{0.888}} & \underline{\textbf{0.817}} & \underline{\textbf{0.888}} \\
16 & NLI & 0.500 & - & \textbf{0.710} & - & 0.221 & - & 0.212 & - & 0.566 & - & \underline{\textbf{0.893}} & - \\
17 & RC & 0.460 & - & \underline{\textbf{0.615}} & - & 0.135 & - & \textbf{0.295} & - & 0.132 & - & 0.264 & - \\
\midrule
18 & No external k. & 0.437 & 0.485 & \underline{\textbf{0.638}} & \underline{\textbf{0.660}} & 0.138 & 0.110 & 0.387 & 0.159 & 0.167 & 0.199 & \textbf{0.400} & \textbf{0.465} \\
19 & Commonsense & \underline{\textbf{0.788}} & 0.698 & 0.752 & \underline{\textbf{0.815}} & 0.613 & 0.364 & 0.624 & 0.356 & 0.735 & 0.470 & \textbf{0.778} & \textbf{0.526} \\
20 & Math formulas & 0.259 & 0.162 & \underline{\textbf{0.661}} & \underline{\textbf{0.544}} & 0.137 & 0.074 & 0.454 & 0.382 & 0.170 & 0.077 & \textbf{0.599} & \textbf{0.404} \\
21 & Science formulas & 0.305 & 0.120 & \underline{\textbf{0.315}} & \underline{\textbf{\textbf{0.250}}} & 0.158 & 0.025 & \textbf{0.239} & 0.021 & 0.157 & 0.105 & 0.181 & \underline{\textbf{\textbf{0.250}}} \\
22 & Computer science k. & 0.262 & 0.128 & \underline{\textbf{0.425}} & \underline{\textbf{0.408}} & 0.151 & 0.137 & 0.147 & 0.134 & \textbf{0.232} & \textbf{0.304} & 0.220 & 0.278 \\
23 & Real-world k. & 0.150 & - & \underline{\textbf{0.472}} & - & 0.012 & - & 0.100 & - & 0.019 & - & \textbf{0.160} & - \\
\midrule
 & \textbf{Average score} & 0.384 & 0.384 & \underline{\textbf{0.604}} & \underline{\textbf{0.586}} & 0.204 & 0.177 & 0.394 & 0.238 & 0.252 & 0.268 & \textbf{0.480} & \textbf{0.448} \\
\bottomrule
\end{tabular}
}
\caption{Evaluations of different baselines across 23 tasks in \lila{}. On most tasks, \textbf{Codex} outperforms all baselines while \textbf{\bhaskara-P} outperforms all fine-tuned baselines. A model usually performs worse on the OOD data set. The \textbf{bold} score refers to the best score among models with the \textit{same supervision} method; the \underline{underlined} score refers to the best score among \textit{all} models. GPT-3 and Codex performance is computed on 100 uniformly distributed examples owing to their cost and usage limit. Fine-tuned model performance is calculated on the full test set.}
\label{tbl:summary}
\end{table*}

\section{Results and Analysis}

A summary of all key results on our \lila benchmark are shown in
Table~\ref{tbl:summary}.
In this section, we will discuss the performance of fine-tuned 2.7B GPT-Neo
models (\S\ref{ssec:expr-general}), performance of models along the 4 categories
of tasks (\S\ref{ssec:expr-partition}) and finally, the few-shot performance of
much larger (${\sim}$175B parameters) models (\S\ref{ssec:expr-fewshot}).

\subsection{Results: Fine-tuned Models}
\label{ssec:expr-general}
\paragraph{Multitasking improves IID performance, robustness, and OOD generalization.}
The multi-tasking model (\bhaskara) substantially improves upon the single task models (Neo).
\bhaskara achieves better average in-domain performance than the 23 individual per-task models 
(0.480 vs.\ 0.394 average score),
suggesting that it leverages cross-task structure not present in a single task's training set. 

\begin{table}
  \centering
  \small
  \begin{tabular}{lcccc} 
    \toprule
    \multirow{2}{*}{\textbf{Dimension}} &  \multicolumn{2}{c}{Neo-A} & \multicolumn{2}{c}{Neo-P} \\
                                        & ~IID & OOD & ~IID & OOD \\ 
                                        \midrule
    Math ability & 0.191 & 0.129 & \textbf{0.445} & \textbf{0.188} \\
    Language & 0.189 & 0.147 & \textbf{0.429} & \textbf{0.246} \\
    Format & 0.246 & 0.382 & \textbf{0.372} & \textbf{0.404} \\
    Knowledge & 0.206 & 0.143 & \textbf{0.331} & \textbf{0.213} \\
    \midrule
    Average & 0.208 & 0.200 & \textbf{0.394} & \textbf{0.263} \\
    \bottomrule
  \end{tabular}
  \caption{
    Multi-task models are able to generalize to unseen tasks in some categories.
    Program output (Neo-P) always outperforms number output (Neo-A).
  }
  \label{tab:cross}
\end{table}

We also find that our multi-task model is robust to the linguistic perturbations we test in \lilarobust.
We did not find any degradation in performance when testing on perturbed IID test examples.
Additionally, multi-task training 
substantially improves out-of-domain generalization (0.448 vs.\ 0.238).
The gap between IID and OOD performance is much smaller for \bhaskara than for
the single task models (Table~\ref{tbl:summary}),
and in one case (format) \bhaskara's OOD performance on held-out tasks 
is better than its IID performance (Table~\ref{tab:cross}).
\lila's multi-task structure opens interesting future directions related to developing improved multitasking techniques, and further understanding its benefits.

Lastly, we do not find any benefit to fine-tuning with instructions. Our best instruction tuned model achieves 0.133 F1, whereas the worst non-instruction-tuned multitask model achieves 0.290.

\paragraph{Program synthesis substantially outperforms answer prediction.} 
Synthesizing the program and evaluating it to get an answer substantially outperforms directly predicting the answer.
For instance, multi-task program synthesis (\bhaskara-P) has an average score of 0.480 while  multi-task answer prediction (\bhaskara-A) scores 0.252.
This means models are often able to generate a program that evaluates to the correct answer, even when the model cannot directly compute the answer.

Program synthesis improves over answer prediction in all math categories except \texttt{Geometry}, with the largest improvements in \texttt{Statistics} and \texttt{Linear Algebra}; see Table \ref{qual-statistics} for examples.
We even see benefits of program synthesis in NLI, a classification-based
task.
\lila's unified problem format decouples synthesis from computation, while opening directions for further study on either aspect.

\paragraph{Models leverage symbolic execution and libraries.}
The gap between program synthesis and answer prediction suggests that the neural language model offloads computations to the symbolic Python runtime that are otherwise difficult to compute directly.
We identify two common cases.
First, the model leverages standard Python as a calculator.
For instance, this pattern is common in the \texttt{basic\_math} and \texttt{mul\_div} categories, which involve evaluating arithmetic expressions;  Table \ref{qual-basic} shows examples.
Second, the model is able to call external libraries  that perform sophisticated computations.
For instance, in statistics the model uses \texttt{scipy.stats.entropy} or \texttt{np.linalg.det} in linear algebra while solving  problems (Table \ref{qual-statistics}).

\paragraph{Models occasionally generate non-executable code.}
Roughly 10\% of \bhaskara's IID programs fail to execute. 
86\% of these are \texttt{SyntaxError}s, which often occur because decoding terminates before finishing the program or the model generates a program of the form `2+3=5', which is invalid Python.
The remaining 14\% of execution failures are less trivial, including \texttt{NameError}s (7\%) and \texttt{TypeError}s (1\%) (see Table \ref{qual-error-name}).


\paragraph{\bhaskara is a good starting point for further fine-tuning}
\documentclass{standalone}

\usepackage{booktabs}

\begin{document}
\begin{table}
  \small
  \centering
  \begin{tabular}{r|rrr|rrr}
\toprule
Data & \multicolumn{3}{c}{Answer (\% F1)} & \multicolumn{3}{c}{Program (\% F1)} \\
 & {Neo} & Multi & $\Delta$ & {Neo} & Multi & $\Delta$ \\
\midrule
100\% & 28.4 & 32.3 & +4.0 & 80.0 & 82.4 & +2.5 \\
40\% & 20.0 & 21.1 & +1.2 & 75.2 & 70.3 & -4.9 \\
20\% & 15.8 & 18.4 & +2.6 & 66.3 & 67.1 & +0.8 \\
\bottomrule
\end{tabular}

  \caption{Here we show the results of fine-tuning both GPT-Neo-2.7B (Neo) and \bhaskara (Multi) on 100\%, 40\%, and 20\% of the held-out data from \lilaood. The Multi almost always outperforms Neo (the $\Delta$ column shows the margin).}
  \label{tab:fine}
\end{table}
\end{document}